\documentclass{ailab}

\usepackage{anyfontsize}
\usepackage{pifont}
\usepackage{dsfont}

\newtcolorbox{AIbox}[2][]{aibox,title=#2,#1}
\usepackage{sidecap}

\usepackage{lipsum}

\definecolor{midnightgreen}{rgb}{0.0, 0.29, 0.33}
\definecolor{deepgreen}{HTML}{055c29}
\definecolor{deeppurple}{HTML}{7030a0}
\definecolor{deepblue}{HTML}{171d91}
\definecolor{brown}{HTML}{843c0c}
\definecolor{shadered}{HTML}{ffe5e5}
\definecolor{shadegreen}{HTML}{e5f7ed}
\definecolor{msftBlack}{RGB}{0,0,0}
\definecolor{lightred}{RGB}{255,163,163}
\definecolor{deepred}{RGB}{153,0,0}

\definecolor{lightblue}{rgb}{0.22,0.45,0.70}
\definecolor{queryrewritter}{HTML}{91B1DA}
\definecolor{resultsummarizer}{HTML}{5680A8}
\definecolor{shadecolor}{rgb}{0.92,0.92,0.92}
\definecolor{LightGray}{HTML}{F0F1F2}
\definecolor{barblue}{RGB}{90,120,180}
\definecolor{barorange}{RGB}{225,124,5}

\newcommand{\placeholder}[1]{\textcolor{cyan}{\{#1\}}}

\tcbuselibrary{listings} 
\newtcblisting{codebox}[2][]{ 
  colback=OliveGreen!5,
  colframe=OliveGreen!50!Black!40,
  boxrule=0.3pt,
  left=2pt, right=2pt, top=1pt, bottom=1pt,
  title=#2, 
  listing only, 
  listing options={ 
    style=tcblatex, 
    language=Python, 
    aboveskip=0pt, 
    belowskip=0pt, 
    basicstyle=\ttfamily\small, 
    breaklines=true, 
  },
  #1 
}

\begin{document}

\title{
\textbf{MedCalc-Eval and MedCalc-Env:}
\\
Advancing Medical Calculation Capabilities of Large Language Models
}

\author{
Kangkun Mao\textsuperscript{1},
Jinru Ding\textsuperscript{1},
Jiayuan Chen\textsuperscript{1},
Mouxiao Bian\textsuperscript{1},
Ruiyao Chen\textsuperscript{1},
Xinwei Peng\textsuperscript{1},
Sijie Ren\textsuperscript{1},
Linyang Li\textsuperscript{1},
Jie Xu\textsuperscript{1$\dagger$}
\\
\textsuperscript{1}Shanghai AI Laboratory;
\textsuperscript{$\dagger$}Correspond to: \{xujie\}@pjlab.org.cn
}

\date{}

\definecolor{mygray}{gray}{.92}
\newcommand{\gray}{\cellcolor{mygray}}
\definecolor{baselinecolor}{rgb}{1, 1, 1}
\newcommand{\baseline}{\cellcolor{baselinecolor}}
\definecolor{ourmethodcolor}{rgb}{0.94, 0.97, 1.0}
\newcommand{\ours}{\cellcolor{ourmethodcolor}}

\maketitle

\thispagestyle{firstpage}
\pagestyle{mynormal}

\begin{abstract}
As large language models (LLMs) become increasingly integrated into the medical domain, existing benchmarks have primarily focused on evaluating their capabilities in question answering and descriptive reasoning. However, real-world clinical practice often relies on quantitative reasoning through medical calculators based on equations and rule-based scoring systems, which are essential for evidence-based decision-making. Current benchmarks such as MedCalc-Bench cover only a limited number of calculation tasks and do not comprehensively reflect LLMs's performance in practical clinical computation scenarios.

To address this gap, we introduce MedCalc-Eval, the largest and most comprehensive benchmark for evaluating LLMs’ capabilities in medical calculations. MedCalc-Eval includes over 700+ distinct clinical calculation tasks, categorized into two major types: equation-based calculations (e.g., Cockcroft-Gault, BMI, BSA) and rule-based scoring systems (e.g., Apgar Score, CHA2DS2-VASc, Glasgow Coma Scale). These tasks span a wide range of clinical specialties, including internal medicine, surgery, pediatrics, critical care, obstetrics and gynecology, emergency medicine, neurology, cardiology, pulmonology, urology, and more, creating a significantly broader and more challenging evaluation setting compared to existing benchmarks.

To further enhance LLM performance in medical computation, we also present MedCalc-Env, a reinforcement learning environment built on the InternBootcamp framework. MedCalc-Env is specifically designed to train LLMs in multi-step clinical reasoning and action planning within interactive settings. Using this environment, we fine-tuned a Qwen2.5-32B model with reinforcement learning, achieving state-of-the-art (SOTA) performance on MedCalc-Eval.

Our evaluation demonstrates that the RL-trained model exhibits markedly improved numerical sensitivity, formula selection accuracy, and reasoning robustness across a wide range of medical calculation tasks. Nonetheless, challenges remain in areas such as unit conversion, multi-condition logic, and context understanding. We hope our work sheds light on the quantitative reasoning gaps in current LLMs and inspires further advancements toward building reliable clinical decision support systems powered by AI. Implementation details with related code and datasets will be updated at
\url{https://github.com/maokangkun/MedCalc-Eval}.
\footnote{This project is ongoing. 
We welcome feedback from the community and will frequently update our work.
}
\end{abstract}

\section{Introduction}
\label{sec: intro}

\begin{figure}[t!]
  \centering
  \includegraphics[width=0.8\linewidth]{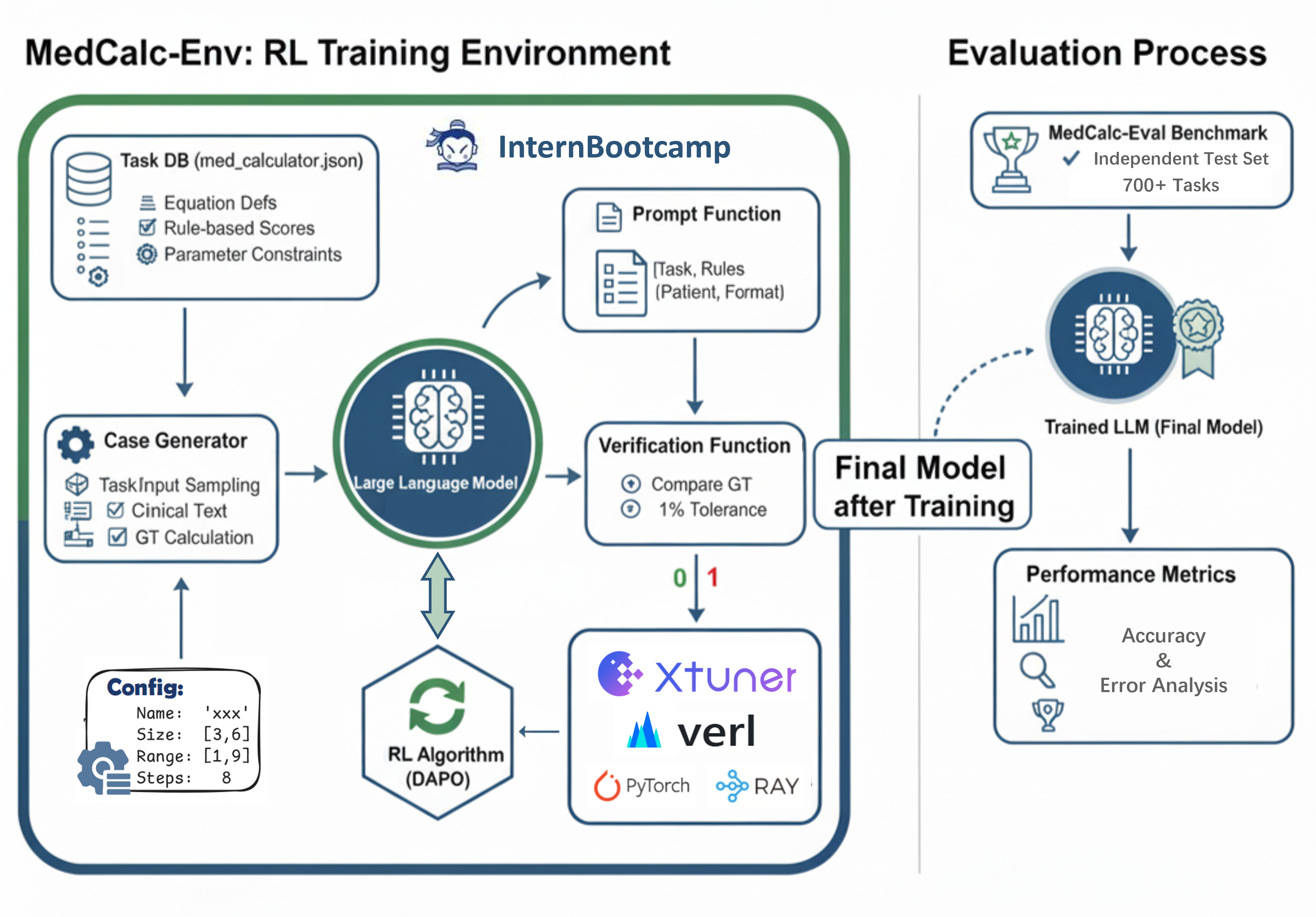}
  \caption{\textbf{Overview of the MedCalc-Env training framework and the MedCalc-Eval evaluation process.} The left side illustrates the reinforcement learning-based MedCalc-Env training loop: (1) The Case Generator samples from the task database to create clinical calculation cases with ground truth answers; (2) The Prompt Function formats the case into an input for the LLM; (3) The LLM generates reasoning steps and a final answer; (4) The Verification Function compares the LLM's answer with the ground truth to generate a reward signal; and (5) The RL algorithm uses this reward signal to update the LLM's model weights. This cycle repeats continuously to enhance the model's capabilities. The right side shows the evaluation process: after full training, the final model is tested on the independent and comprehensive MedCalc-Eval benchmark to objectively measure its final performance and generalization ability on medical calculation tasks.}
  \label{fig: framework}
  \vspace{-8pt}
\end{figure}

The rapid advancements in large language models (LLMs) have led to their increasing integration into various specialized domains, including medicine. These models have demonstrated remarkable capabilities in tasks such as medical question answering, information extraction, and descriptive reasoning. However, the existing benchmarks predominantly focus on these qualitative aspects, often overlooking the critical need for quantitative reasoning in real-world clinical practice. Medical professionals frequently rely on clinical calculators, which are built upon precise equations and rule-based scoring systems, to make evidence-based decisions. These tools are indispensable for accurate diagnosis, prognosis, and treatment planning.

Despite the importance of quantitative reasoning, current evaluation benchmarks for LLMs in medicine, such as MedCalc-Bench~\citep{khandekar2023medcalc}, offer a limited scope. While MedCalc-Bench was a pioneering effort to introduce clinical calculation scenarios, it covers only a restricted number of calculation tasks and does not fully capture the complexities and breadth of practical clinical computation. This gap highlights a significant challenge: without robust quantitative reasoning capabilities, LLMs cannot fully support the intricate decision-making processes inherent in medical practice.

To address this critical limitation, we introduce \textbf{MedCalc-Eval}, a novel and comprehensive benchmark designed specifically for evaluating the medical calculation capabilities of LLMs. MedCalc-Eval significantly expands upon previous efforts by encompassing over 700 distinct clinical calculation tasks. These tasks are meticulously categorized into two primary types: equation-based calculations, which involve direct mathematical formulas (e.g., Cockcroft-Gault formula for creatinine clearance, Body Mass Index (BMI), Body Surface Area (BSA)), and rule-based scoring systems, which require logical inference based on predefined criteria (e.g., Apgar Score for newborn assessment, CHA2DS2-VASc score for stroke risk in atrial fibrillation, Glasgow Coma Scale for assessing consciousness). The benchmark spans a wide array of clinical specialties, including internal medicine, surgery, pediatrics, critical care, obstetrics and gynecology, emergency medicine, neurology, cardiology, pulmonology, urology, and many others. This broad coverage creates a significantly broader and more challenging evaluation setting compared to existing benchmarks, providing a more accurate assessment of LLMs's performance in diverse clinical computation scenarios.

Furthermore, to facilitate the enhancement of LLM performance in medical computation, we also present \textbf{MedCalc-Env}, a reinforcement learning environment. Built upon the robust InternBootcamp framework, MedCalc-Env is specifically engineered to train LLMs in multi-step clinical reasoning and action planning within interactive settings. This environment allows models to learn from iterative interactions and receive immediate feedback, thereby refining their ability to handle complex medical calculation tasks. Through the utilization of MedCalc-Env, we successfully fine-tuned a Qwen2.5-32B model using reinforcement learning techniques. This RL-trained model achieved state-of-the-art (SOTA) performance on both our newly introduced MedCalc-Eval benchmark and the existing MedCalc-Bench, demonstrating the effectiveness of our approach in improving quantitative reasoning in LLMs.

Our comprehensive evaluation reveals that the reinforcement learning-trained model exhibits significantly improved numerical sensitivity, enhanced formula selection accuracy, and greater reasoning robustness across a broad spectrum of medical calculation tasks. However, our analysis also identifies persistent challenges, particularly in areas such as unit conversion, multi-condition logic, and nuanced context understanding. These findings underscore the remaining quantitative reasoning gaps in current LLMs and highlight crucial areas for future research. We believe that our work not only provides a more rigorous evaluation framework but also inspires further advancements towards the development of highly reliable and intelligent clinical decision support systems powered by AI. (Figure~\ref{fig: framework})

Our contributions can be summarized as follows:
\begin{itemize}
    \item We introduce \textbf{MedCalc-Eval}, the largest and most comprehensive benchmark for evaluating LLMs's medical calculation capabilities, covering over 700+ tasks across various specialties.
    \item We develop \textbf{MedCalc-Env}, a reinforcement learning environment based on InternBootcamp, designed to train LLMs in multi-step clinical reasoning for medical computations.
    \item We demonstrate that an RL-trained Qwen2.5-32B model achieves state-of-the-art performance on both MedCalc-Eval and MedCalc-Bench, showcasing the effectiveness of our proposed training methodology.
    \item We provide a detailed error analysis, identifying key challenges and future research directions for improving quantitative reasoning in medical LLMs.
\end{itemize}

\section{MedCalc-Eval: A Comprehensive Evaluation Benchmark}
\label{sec: framework}

To overcome the limitations of existing benchmarks and provide a more rigorous evaluation of LLMs's medical calculation capabilities, we propose \textbf{MedCalc-Eval}. This benchmark is designed to be the largest and most comprehensive of its kind, encompassing a vast array of clinical calculation tasks that closely mirror real-world medical practice. MedCalc-Eval significantly expands the scope and depth of evaluation, offering a more accurate assessment of LLMs's performance in diverse and challenging clinical scenarios.

\subsection{Task Definition}

Medical clinical calculator tasks, in the context of clinical diagnosis and treatment, involve a quantitative reasoning process based on patient medical record information and established medical formulas or scoring rules. These tasks typically comprise three critical stages:
\begin{enumerate}
    \item \textbf{Knowledge Recall}: Correctly identifying and invoking the appropriate clinical formula or scoring scale relevant to the given medical scenario.
    \item \textbf{Information Extraction}: Accurately extracting relevant parameters (numerical values, categories, time points, etc.) from lengthy or complex medical record texts. This often requires sophisticated natural language understanding to identify key data points amidst noise.
    \item \textbf{Numerical Reasoning}: Performing multi-step calculations and logical judgments based on correctly substituted parameters, and outputting the final result. This stage demands precision, adherence to specific rules, and often involves complex arithmetic or conditional logic.
\end{enumerate}
Unlike traditional open-domain question answering or descriptive reasoning tasks, medical clinical calculator tasks are highly structured, emphasizing precision, compliance, and verifiability. They form a crucial component of Clinical Decision Support Systems (CDSS) and serve as a vital indicator for assessing the practical applicability of LLMs in medical settings.

\subsection{Construction of MedCalc-Eval}

MedCalc-Eval is meticulously constructed to provide a comprehensive and challenging evaluation environment. As summarized in Table~\ref{tab:dataset_summary_stats} (see Appendix~\ref{detail: eval}), the benchmark contains a total of \textbf{709 distinct clinical calculation tasks} (629 formula-based and 80 scale-based), making it the largest collection of its kind. Its key characteristics include:
\begin{itemize}
    \item \textbf{Task Scale and Diversity}: MedCalc-Eval includes \textbf{132 formula-based categories} and \textbf{27 scale-based categories}, covering a wide range of clinical scenarios. This extensive scale ensures a broad coverage of medical calculation scenarios.
    \item \textbf{Task Types}: The benchmark categorizes tasks into two major types:
    \begin{itemize}
        \item \textbf{Equation-based calculations}: These involve direct application of mathematical formulas, such as the Cockcroft-Gault formula for creatinine clearance, Body Mass Index (BMI), and Body Surface Area (BSA).
        \item \textbf{Rule-based scoring systems}: These require logical inference and scoring based on predefined criteria, exemplified by the Apgar Score, CHA2DS2-VASc score, and Glasgow Coma Scale.
    \end{itemize}
    \item \textbf{Specialty Coverage}: MedCalc-Eval spans dozens of clinical specialties. As shown in Figure~\ref{fig:medcalc_eval_categories} and detailed in Table~\ref{tab:combined_top20} (see Appendix~\ref{detail: eval}), the top formula-based categories include Laboratory Medicine, Pulmonary Diseases, and Nephrology, while scale-based questions frequently appear in Cardiovascular Diseases and Obstetrics and Gynecology. This broad specialty coverage ensures that the benchmark reflects the diverse needs of real-world clinical practice.
    \item \textbf{Increased Difficulty}: To provide a more realistic and challenging evaluation, MedCalc-Eval incorporates scenarios that involve multi-condition judgments, complex formula nesting, and cross-unit conversions. The most frequently required input indicators, such as \textbf{Weight}, \textbf{Age}, and \textbf{Height}, often require careful unit handling, as detailed in Table~\ref{tab:top_formula_indicators} (see Appendix~\ref{detail: eval}).
\end{itemize}

\begin{figure}[htbp]
    \centering
    \includegraphics[width=\linewidth]{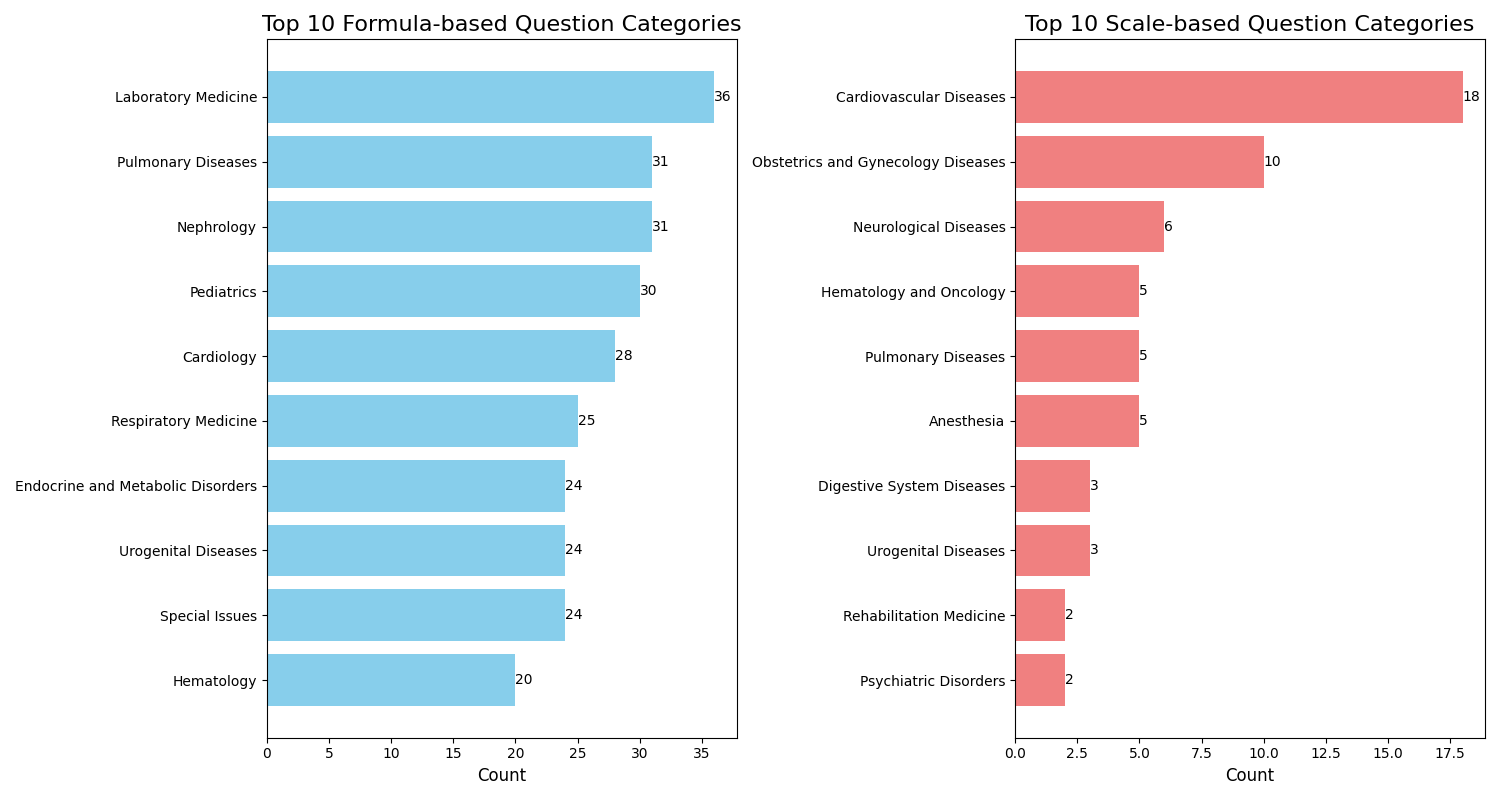}
    \caption{Top 10 categories in MedCalc-Eval}
    \label{fig:medcalc_eval_categories}
\end{figure}

\subsection{Comparison with MedCalc-Bench}

While MedCalc-Bench~\citep{khandekar2023medcalc} laid the groundwork for evaluating LLMs in medical calculation, MedCalc-Eval represents a significant advancement in both breadth and depth. MedCalc-Bench covered 55 calculators and 1047 instances, with specific error tolerances for different task types. However, its limitations included a relatively small number of calculation tasks, insufficient specialty coverage, and a lack of challenging scenarios involving complex reasoning chains, multi-modal inputs, or cross-unit conversions. These limitations suggest that while MedCalc-Bench laid a foundational framework, it does not fully characterize the true capabilities and boundaries of LLMs in complex medical calculation tasks.

MedCalc-Eval directly addresses these shortcomings by:
\begin{itemize}
    \item \textbf{Vastly expanded task variety}: Offering \textbf{709 distinct tasks} compared to MedCalc-Bench's 55, providing a much richer and more diverse evaluation landscape.
    \item \textbf{Comprehensive specialty representation}: Covering a wider range of clinical specialties, ensuring that LLMs are tested across the full spectrum of medical domains.
    \item \textbf{Enhanced complexity}: Introducing more challenging scenarios that demand advanced reasoning, such as nested formulas, conditional logic, and unit conversions, which are crucial for practical clinical application.
\end{itemize}
In summary, MedCalc-Eval significantly extends the evaluation system's breadth and depth, providing a more stringent and systematic test for the medical calculation capabilities of LLMs. It serves as a more robust platform for future research and development in this critical area.

\section{MedCalc-Env: A Reinforcement Learning Environment}
\label{sec: env}

Beyond the comprehensive evaluation benchmark, we have also developed \textbf{MedCalc-Env}, a novel reinforcement learning (RL) environment built upon the InternBootcamp framework. MedCalc-Env is specifically designed to train large language models in multi-step clinical reasoning and action planning within interactive settings, thereby enhancing their quantitative reasoning capabilities in medical computation tasks. The overall architecture of MedCalc-Env is illustrated in Appendix~\ref{detail: code}.

\subsection{Case Generator}

The case generator in MedCalc-Env is crucial for producing diverse and realistic clinical scenarios, ensuring both input randomness and the verifiability of output results. The overall logic of the case generator is illustrated in the following. It operates by sampling tasks from two main categories: {equation} (formula-based) and {scale} (rule-based), using the detailed configuration JSON file.

\subsubsection{Overall Process}

The process begins with \textbf{Task Sampling}, where a category (either equation or scale) is randomly selected, followed by the random selection of a specific calculator or scoring table within that category. Subsequently, the \texttt{gen\_a\_case} method is invoked to generate a complete instance based on the formula definitions, indicator constraints, and scale items defined in the configuration file (e.g., \texttt{med\_calculator.json}).

\subsubsection{Formula-based Tasks}

For formula-based tasks, the system performs \textbf{Input Sampling} by iterating through the required input parameters for a given formula. It reads the type and range from the {indicator} dictionary in the configuration file:
\begin{itemize}
    \item \textbf{Integer (int)}: A random integer is selected within the specified range.
    \item \textbf{Float (float)}: A random float value is generated within the range, with precision maintained as per constraints.
    \item \textbf{Choice (choice)}: A random item is chosen from a given set of options (list or dictionary), returning the mapped numerical value if it's a dictionary.
\end{itemize}
These sampled parameters are then transformed into \textbf{Clinical Expression}, concatenating parameter values with units to form patient information descriptions (e.g., "blood pressure 120 mmHg"). The sampled results are then substituted into the original formula for \textbf{Formula Substitution and Solving}. The system uses Python's {math} library and {eval} function to compute the result. Error handling is implemented to re-sample if invalid values (e.g., division by zero) occur. Finally, \textbf{Output Standardization} ensures the target value is rounded as per the result indicator's configuration.

\subsubsection{Scale-based Tasks}

For scale-based tasks, the system iterates through all items in the scale. For \textbf{Single-choice} items, one option is randomly selected, and its corresponding score is accumulated. For \textbf{Multi-choice} items, a random number of options are selected, and their scores are summed. These selections are then used to generate \textbf{Input Descriptions} that mimic real medical records (e.g., "symptoms: chest pain, shortness of breath..."). The \textbf{Score Calculation} sums the scores of selected options to derive the final target score.

\subsubsection{Randomization and Robustness}

MedCalc-Env incorporates several mechanisms to ensure the diversity and robustness of generated cases:
\begin{itemize}
    \item \textbf{Multi-source Randomness}: Including task categories, specific calculators, input values/options, and the inclusion of rule descriptions, ensuring diverse data distribution.
    \item \textbf{Error Tolerance}: Catching and re-sampling for errors like division by zero or invalid values, ensuring all generated results are valid.
    \item \textbf{Enhanced Realism}: Automatically concatenating units and labels in input descriptions to conform to clinical medical record conventions.
    \item \textbf{Random Inclusion of Rules}: Each case has a probability (controlled by the \texttt{add\_rule\_ratio} parameter) to include the formula explanation or scale scoring criteria in the prompt. This feature allows for evaluating the model's performance in both scenarios: when it must rely solely on its internal medical knowledge versus when explicit calculation rules are provided.
\end{itemize}

\subsection{Prompt Function Setting}

The prompt function serves as the "interface contract" connecting the large language model with the verification environment. It structures the task intent, input data, and constraints into executable instructions, and constrains the model's output to a machine-readable and verifiable format. Two main variations of prompt templates are designed based on whether knowledge and rules are provided. The detailed prompt templates are provided in Appendix~\ref{detail: prompt}.

\subsection{Reinforcement Learning with Verifiable Rewards (RLVR)}

For our training paradigm, we use \textbf{Reinforcement Learning with Verifiable Rewards (RLVR)}~\citep{wen2025reinforcementlearningverifiablerewards}, a method that implicitly incentivizes correct reasoning in the base LLM by providing a reward signal based on the verifiable final output. The core idea is to leverage the deterministic nature of medical calculation tasks to create a strong, objective reward signal.

The verification function determines the reward signal:
\begin{itemize}
    \item \textbf{Reward = 1}: If the model's final answer, extracted from the boxed output, matches the ground truth value calculated by the simulation tool.
    \item \textbf{Reward = 0}: Otherwise.
\end{itemize}
To account for potential minor numerical discrepancies inherent in complex, multi-step calculations and floating-point arithmetic, we introduce an error tolerance for formula-based tasks (e.g., laboratory/physical/dosage conversions). Specifically, an error tolerance of $\pm 1\%$ is allowed, meaning if the difference between the model's predicted result and the ground truth falls within this range, it is still considered correct. This approach facilitates more robust training by focusing on the correctness of the final numerical result, which is the ultimate verifiable metric in these tasks. This verifiable reward mechanism, by focusing the reward signal solely on the final verifiable output, implicitly incentivizes the model to generate correct, multi-step reasoning, thereby addressing extraction and calculation errors.

\section{Experiments}
\label{sec: exp}

This section details the experimental setup, presents the results obtained from evaluating our RL-trained model on MedCalc-Eval and MedCalc-Bench, and provides an in-depth analysis of the error types observed.

\subsection{Experimental Setup}

Our experimental validation involved leveraging the MedCalc-Env to train models using the Reinforcement Learning with Verifiable Rewards (RLVR) methodology~\citep{wen2025reinforcementlearningverifiablerewards}, demonstrating its effectiveness in enhancing quantitative reasoning.

\subsubsection{Model Selection \& Training Workflow}
We employed the \textbf{Qwen2.5-7B} and \textbf{Qwen2.5-32B}~\citep{qwen2025qwen25technicalreport} model for our Reinforcement Learning with Verifiable Rewards (RLVR) experiments, which was fine-tuned within the MedCalc-Env environment. We also evaluated several other large language models~\citep{qwen2025qwen25technicalreport,qwen3,deepseekai2025deepseekr1incentivizingreasoningcapability} in a zero-shot setting for comparison.
As for training process, we total generated 10k corresponding data instances for all formula and scale in the MedCalc-Env bootcamp to construct the training environment for RLVR.
During the RLVR stage, the following key training parameters were configured:
\begin{itemize}
    \item \textbf{Algorithm}: Dynamic sAmpling Policy Optimization (DAPO)~\citep{yu2025dapoopensourcellmreinforcement}, with No KL-loss.
    \item \textbf{Training Prompt Batch Size}: 128.
    \item \textbf{Maximum Response Length}: 8192.
    \item \textbf{Temperature}: 1.
    \item \textbf{Rollout Number}: 8.
    \item \textbf{Loss Aggregation Mode}: token-mean.
\end{itemize}
We utilized \textbf{xpuyu-RL}~\citep{2023xtuner} and \textbf{VeRL}~\citep{Sheng_2025} frameworks for the training process.

\subsubsection{Experimental Data}

\begin{itemize}
    \item \textbf{Training Data}: The training dataset was exclusively generated using the MedCalc-Env bootcamp, ensuring consistency and relevance to the medical calculation tasks.
    \item \textbf{Testing Data}: For evaluation, we used a separate test set generated by the MedCalc-Env bootcamp and subsequently verified manually, ensuring high quality and reliability of the evaluation.
\end{itemize}

\subsection{Experimental Results and Analysis}

Our experiments yielded significant improvements in the LLMs's medical calculation capabilities, demonstrating the effectiveness of the proposed MedCalc-Env training environment.

\subsubsection{Main Results}

\begin{figure}[htbp]
    \centering
    \includegraphics[width=\linewidth]{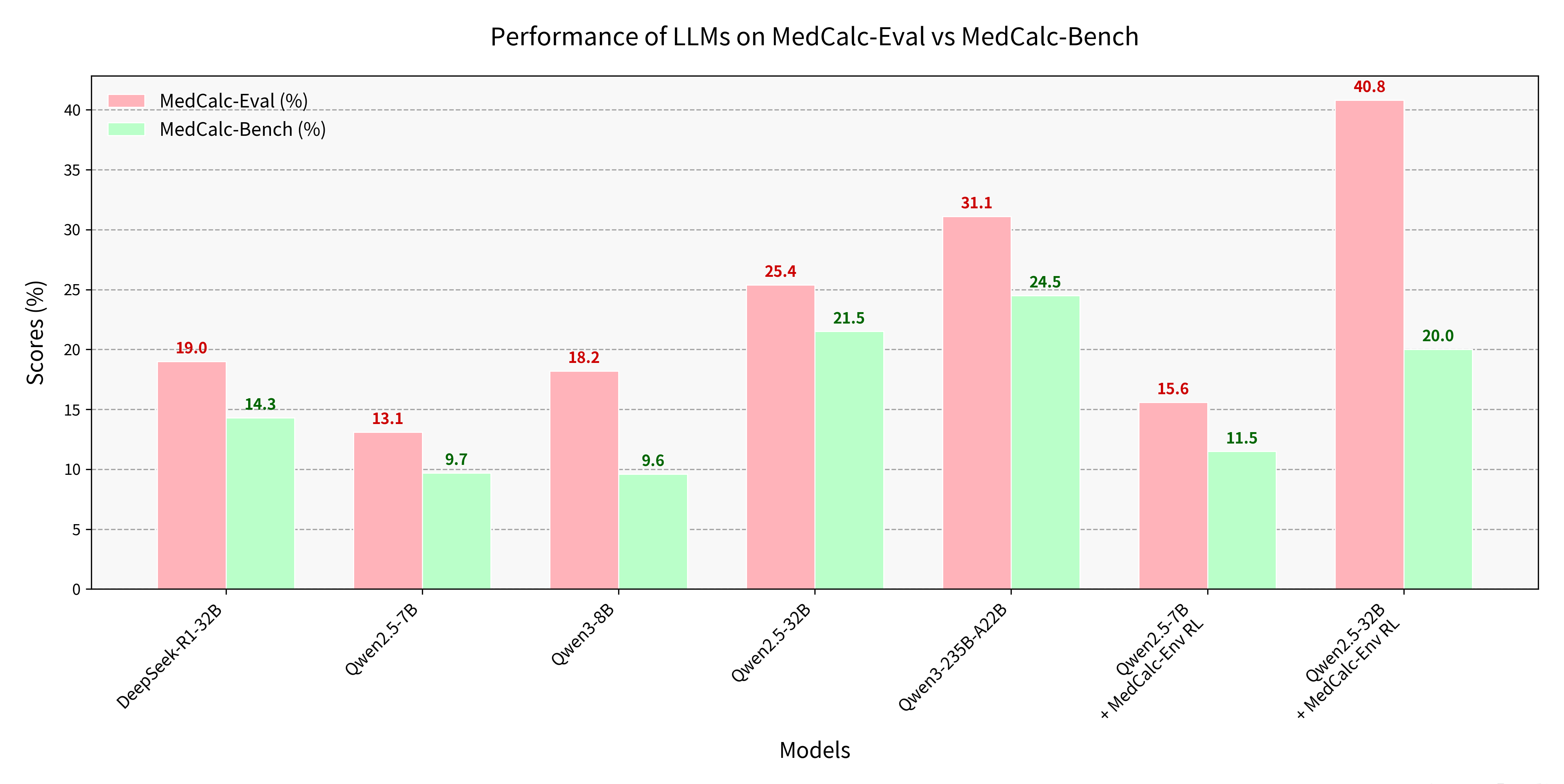}
    \caption{Performance of LLMs on MedCalc-Eval compared to MedCalc-Bench}
    \label{fig:main-results}
\end{figure}

Figure~\ref{fig:main-results} illustrates the performance comparison of various large language models on both the MedCalc-Eval and MedCalc-Bench benchmarks. The detailed data is provided in Appendix~\ref{tab:main_results}.

The results show that:
\begin{itemize}
    \item The performance of all evaluated large language models is generally poor on both \textbf{MedCalc-Eval} and \textbf{MedCalc-Bench}. The accuracy of most models is below 30\%, with the best zero-shot performance from \textbf{Qwen3-235B-A22B} only reaching 31.1\% on MedCalc-Eval. This collectively demonstrates that existing LLMs still have significant shortcomings in medical calculation tasks, indicating a large room for improvement. Furthermore, MedCalc-Eval (Chinese medical calculation) and MedCalc-Bench (English medical calculation) serve as complementary benchmarks for comprehensively evaluating the medical calculation capabilities of LLMs.
    \item The Reinforcement Learning (RL) approach trained within our \textbf{MedCalc-Env} environment leads to substantial performance gains. Specifically, the \textbf{Qwen2.5-32B + MedCalc-Env RL} model achieved the highest accuracy of \textbf{40.8\%} on MedCalc-Eval, representing an absolute improvement of \textbf{15.4\%} over its base model (25.4\%). This substantial gain underscores the effectiveness of our proposed RL training environment in enhancing quantitative reasoning for medical tasks.
    \item The largest model, \textbf{Qwen3-235B-A22B}, shows the highest zero-shot performance (31.1\%), but the RL-trained Qwen2.5-32B model surpasses it by a large margin (40.8\%), indicating that specialized training with MedCalc-Env is more effective than simply scaling up model size.
\end{itemize}

\subsubsection{Ablation Study}

To further validate the effectiveness of the MedCalc-Env training, we conducted an ablation study using the Qwen2.5-7B model, as summarized in Appendix~\ref{tab:abalation_results}.
\begin{table}[htbp]
\centering
\caption{Ablation study results of Qwen2.5-7B model}
\begin{tabular}{|l|c|c|c|}
\hline
\textbf{Model} & \textbf{MedCalc-Eval (\%)} & \textbf{MedCalc-Bench (\%)} & \textbf{AIME24 (\%)} \\
\hline
Qwen2.5-7B & 13.1 & 9.7 & 16.7 \\
\hline
Qwen2.5-7B + MedCalc-Env RL & 15.6 & 11.5 & 20.0 \\
\hline
\end{tabular}
\label{tab:abalation_results}
\end{table}

The ablation study confirms the positive impact of the RL training:
\begin{itemize}
    \item The \textbf{Qwen2.5-7B + MedCalc-Env RL} model shows a consistent improvement across all three evaluation sets: MedCalc-Eval (13.1\% to 15.6\%), MedCalc-Bench (9.7\% to 11.5\%), and the general mathematical evaluation AIME24 (16.7\% to 20.0\%).
    \item Notably, the model trained primarily with Chinese medical data in the MedCalc-Env environment also achieved performance improvement on the English medical calculation benchmark MedCalc-Bench (from 9.7\%to 11.5\%), demonstrating that the performance gain stems from an enhancement in the model's fundamental medical calculation capabilities, which are largely language-agnostic.
    \item The improvement on AIME24 (an increase of 3.3\% absolute, or 20\% relative) is particularly noteworthy, as it was trained with a mixed dataset comprising 10,000 medical data instances and 17,000 mathematical data instances (DAPO\_MATH\_17K). This result validates that integrating specialized medical knowledge through our RL environment does not lead to a loss of generalization performance in other domains, but rather \textbf{enhances the model's core quantitative reasoning capabilities}.
\end{itemize}

\subsubsection{Case Study \& Error Analysis}

To gain deeper insights into the model's reasoning process and the nature of its errors, we conducted detailed case studies of both scale-based and formula-based medical calculations. Appendix~\ref{fig: full output for scale based question} and Appendix~\ref{fig: full output for formula based question} present complete model responses for CURB-65 pneumonia severity scoring and serum osmolality calculation, respectively.

These case studies reveal that our RL-trained model demonstrates strong multi-step reasoning capabilities: it correctly identifies relevant clinical parameters, applies appropriate medical formulas or scoring criteria, and executes precise numerical computations. The model's ability to maintain coherent reasoning chains while handling complex unit conversions and numerical precision requirements highlights the effectiveness of the MedCalc-Env training.

Based on our observations and the analysis presented in the supplementary material, we categorize the errors made by LLMs into four main types:
\begin{itemize}
    \item \textbf{Type A: Knowledge Errors}: These occur when the model incorrectly recalls or applies a formula or rule. In zero-shot settings, this category typically accounts for the highest proportion of errors.
    \item \textbf{Type B: Extraction Errors}: These errors arise when the model fails to correctly identify or extract attribute values from the patient's medical record. This often involves issues with natural language understanding, synonym recognition, or discerning relevant information from noisy contexts.
    \item \textbf{Type C: Calculation Errors}: These are arithmetic errors where the model correctly identifies the formula and extracts the values but makes mistakes during the numerical computation. This highlights a vulnerability in the execution phase of the calculation task.
    \item \textbf{Type D: Other Errors}: This category includes miscellaneous errors such as improper formatting, incorrect unit conversions, or non-standard date outputs.
\end{itemize}

Our comprehensive error analysis across the evaluation datasets indicates a significant shift in error distribution patterns. In zero-shot conditions, Type A (knowledge errors) constituted the majority of failures, indicating fundamental gaps in medical knowledge recall. However, after targeted interventions including medical knowledge prompting and especially after MedCalc-Env RL training, we observed a dramatic reduction in Type A errors, while Type B (extraction errors) and Type C (calculation errors) also decreased.

The MedCalc-Env RL training directly addresses these remaining challenges. By emphasizing accurate attribute extraction, medical knowledge recall, proper formula application, and error-free numerical computation, our approach specifically targets these bottlenecks, ultimately leading to more reliable and trustworthy medical calculation capabilities.

\section{Related Work}
\label{sec: related work}

\subsection{LLMs in Medical Domain and Qualitative Benchmarks}
The application of Large Language Models (LLMs) in medicine has seen rapid growth~\citep{wang2025baichuanm1pushingmedicalcapability,xie2024llamafoundationlargelanguage,labrak2024biomistralcollectionopensourcepretrained,sallinen2025llama3_meditron,wu2025medreasonelicitingfactualmedical,liu2024medcotmedicalchainthought,lai2025med}, primarily focusing on tasks that rely on natural language understanding and generation. Early and foundational benchmarks, such as \textbf{MedQA}~\citep{medqa}, \textbf{PubMedQA}~\citep{pubmedqa}, and \textbf{MedMCQA}~\citep{medmcqa}, have been crucial in assessing LLMs's medical knowledge, diagnostic capabilities, and ability to follow clinical guidelines. These benchmarks typically involve multiple-choice questions, open-ended descriptive reasoning, and information retrieval, which are essential for evaluating the qualitative aspects of medical AI. More recent and holistic evaluation frameworks, like \textbf{HealthBench}~\citep{arora2025healthbenchevaluatinglargelanguage} and the \textbf{Swedish Medical LLM Benchmark (SMLB)}~\citep{moell2025swedish}, aim to provide a more comprehensive assessment, including safety and ethical considerations. However, these efforts generally emphasize descriptive and diagnostic reasoning, often overlooking the critical need for precise quantitative skills in clinical practice.

\subsection{Quantitative Reasoning and Medical Calculation Benchmarks}
Quantitative reasoning, which involves precise numerical computation and logical application of formulas and rules, is indispensable for evidence-based medical decision-making. Clinical calculators, built upon established equations and rule-based scoring systems, are routinely used for risk assessment, drug dosage calculation, and physiological parameter estimation.
\textbf{MedCalc-Bench}~\citep{khandekar2023medcalc}, introduced as the first systematic benchmark for clinical calculation scenarios, marked a significant step towards evaluating LLMs's capabilities in this area. It includes over 50 different medical calculators and provides a foundational framework for assessing numerical accuracy.
However, as discussed in our introduction, MedCalc-Bench's coverage is limited, lacking comprehensive representation across various medical specialties and complex, multi-step calculation tasks. Recent works, such as \textbf{MedRaC}~\citep{wang2025scoresstepsdiagnosingimproving}, have focused on diagnosing and improving LLM performance in evidence-based medical calculations, highlighting the ongoing challenge of numerical sensitivity and reasoning chain robustness. Our \textbf{MedCalc-Eval} significantly expands this landscape by offering a substantially larger and more diverse set of calculation tasks, including a wider range of specialties and more complex rule-based systems, providing a more rigorous stress test for LLMs's quantitative reasoning.

\subsection{Reinforcement Learning for Enhanced LLM Reasoning}
The challenge of multi-step and complex reasoning in LLMs has led to the exploration of advanced training paradigms, particularly those involving interactive learning and feedback. The \textbf{InternBootcamp}~\citep{li2025internbootcamptechnicalreportboosting} framework, on which our training environment is based, represents a novel approach to continuous model evolution through interactive feedback. This closed-loop system allows models to acquire and refine complex reasoning abilities by integrating evaluation directly into the training process.
In the medical domain, Reinforcement Learning (RL) is increasingly being leveraged to align LLMs with complex clinical protocols and improve reasoning. Works like \textbf{EHRMIND}~\citep{lin2025trainingllmsehrbasedreasoning} and \textbf{Baichuan-M2}~\citep{m2team2025baichuanm2scalingmedicalcapability} demonstrate the application of RL to tasks like EHR-based reasoning and the emergence of reasoning capability through RL, respectively. Furthermore, the use of RL for structured medical reasoning, such as aligning with ACR Imaging Appropriateness Criteria, underscores the potential of this approach. Our \textbf{MedCalc-Env} builds upon this foundation, adapting the interactive and feedback-driven nature of InternBootcamp and RL to the specific challenges of medical calculation, enabling the model to learn robust formula selection, unit handling, and multi-step computational accuracy.

\section{Conclusion}

In this work, we have successfully integrated the medical clinical calculator question-answering task into general-purpose large language models through the InternBootcamp framework. Our experimental results unequivocally demonstrate that models trained using the InternBootcamp approach exhibit significantly enhanced capabilities in medical clinical calculator tasks. This further validates the feasibility and effectiveness of the "general-purpose and specialized integration" technical route, providing a reusable methodology and valuable reference for future practices in medical domain-specific integration.

We introduced \textbf{MedCalc-Eval}, a comprehensive and challenging benchmark that addresses the limitations of existing evaluation datasets by offering a significantly expanded scope of medical calculation tasks across numerous clinical specialties. Complementing this, we developed \textbf{MedCalc-Env}, a reinforcement learning environment designed to foster multi-step clinical reasoning and action planning in LLMs. The state-of-the-art performance achieved by our RL-trained Qwen2.5-32B model on both MedCalc-Eval and the existing MedCalc-Bench underscores the potential of our approach to bridge the quantitative reasoning gap in current LLMs.

Despite these advancements, our error analysis highlights several areas for future improvement. Challenges persist in tasks requiring precise unit conversion, handling complex multi-condition logic, and achieving nuanced context understanding from patient notes. These areas represent critical avenues for future research to further enhance the reliability and accuracy of AI-powered clinical decision support systems.

For future work, we envision several directions:
\begin{itemize}
    \item \textbf{Dynamic, Realistic Patient Simulation}: A critical next step is to move beyond static, structured benchmark data towards dynamically generating realistic, free-text patient descriptions as model inputs. This involves creating simulated clinical narratives (e.g., physician progress notes, emergency department reports) where patient attributes, medical histories, and clinical findings are presented in natural, unstructured language, complete with redundancies, synonyms, and irrelevant information. Training and evaluating models on such data will significantly enhance their robustness in information extraction and their applicability to real-world clinical text.
    \item \textbf{Multi-modal Integration}: Exploring the integration of multi-modal inputs (e.g., medical images, physiological signals) to enable more comprehensive and context-aware medical calculations.
    \item \textbf{Explainable AI (XAI)}: Developing methods to make the LLM's reasoning process for medical calculations more transparent and explainable, which is crucial for trust and adoption in clinical settings, especially when handling complex, narrative-style inputs.
    \item \textbf{Cross-lingual and Cross-cultural Generalization}: Extending the model's capabilities to handle medical calculations across different languages, clinical guidelines, and healthcare systems, improving global applicability.
\end{itemize}
We believe that continued research in these areas will pave the way for more robust, accurate, and clinically applicable large language models, ultimately contributing to safer and more efficient healthcare delivery.

\bibliography{ref}
\bibliographystyle{ref-style}

\clearpage
\onecolumn
\appendix 
\etocdepthtag.toc{mtappendix}
\etocsettagdepth{mtchapter}{none}
\etocsettagdepth{mtappendix}{subsection}
\renewcommand{\contentsname}{Appendix}
\tableofcontents
\clearpage

\section{Details of MedCalc-Eval}
\label{detail: eval}

In this section, we show the details of our MedCalc-Eval.

\subsection{Dataset Details}
\begin{table}[htbp]
    \centering
    \caption{Summary Statistics of the MedCalc-Eval}
    \label{tab:dataset_summary_stats}
    \begin{tabular}{l c c c}
        \toprule
         & \textbf{Formula-based} & \textbf{Scale-based} & \textbf{Indicators} \\
        \midrule
        \textbf{Categories} & 132 & 27 & - \\
        \textbf{Total Numbers} & 629 & 80 & 1432 \\
        \bottomrule
    \end{tabular}
\end{table}

\subsubsection{Top 20 Categories in Formula-based or Scale-based Question}
\begin{table*}[htbp] 
    \centering
    \caption{Top 20 Categories in Formula-based or Scale-based Question}
    \label{tab:combined_top20}
    
    \begin{minipage}[t]{0.49\textwidth}
        \centering
        \subcaption{Formula-based Question}
        \label{tab:formula_category_top20_en}
        \begin{tabular}{r l r}
            \toprule
            Rank & Category & Count \\
            \midrule
            1 & Laboratory Medicine & 36 \\
            2 & Pulmonary Diseases & 31 \\
            3 & Nephrology & 31 \\
            4 & Pediatrics & 30 \\
            5 & Cardiology & 28 \\
            6 & Respiratory Medicine & 25 \\
            7 & Special Issues & 24 \\
            8 & Urogenital Diseases & 24 \\
            9 & Endocrine and Metabolic Disorders & 24 \\
            10 & Hematology & 20 \\
            11 & Respiratory Medicine/ICU & 16 \\
            12 & Emergency Department & 16 \\
            13 & Pharmacy & 15 \\
            14 & Radiology & 14 \\
            15 & Internal Medicine & 14 \\
            16 & Emergency Medicine & 13 \\
            17 & Pharmacology & 12 \\
            18 & Hepatobiliary Diseases & 11 \\
            19 & Anesthesiology & 10 \\
            20 & Nutritional Diseases & 9 \\
            \bottomrule
        \end{tabular}
    \end{minipage}
    \hfill 
    \begin{minipage}[t]{0.45\textwidth}
        \centering
        \subcaption{Scale-based Question}
        \label{tab:scale_category_top20_en}
        \begin{tabular}{l r}
            \toprule
            Category & Count \\
            \midrule
            Cardiovascular Diseases & 18 \\
            Obstetrics and Gynecology Diseases & 10 \\
            Neurological Diseases & 6 \\
            Hematology and Oncology & 5 \\
            Pulmonary Diseases & 5 \\
            Anesthesia & 5 \\
            Urogenital Diseases & 3 \\
            Digestive System Diseases & 3 \\
            Rehabilitation Medicine & 2 \\
            Hepatobiliary Diseases & 2 \\
            Pediatric Science & 2 \\
            Otorhinolaryngology & 2 \\
            Psychiatric Disorders & 2 \\
            Respiratory Medicine & 2 \\
            Emergency Medicine & 1 \\
            Special Issues & 1 \\
            Gastrointestinal Diseases & 1 \\
            Geriatrics & 1 \\
            Emergency Department & 1 \\
            Intensive Care Unit & 1 \\
            \bottomrule
        \end{tabular}
    \end{minipage}
\end{table*}

\subsubsection{Formula Input Indicators}
\begin{table}[htbp]
    \centering
    \caption{Most Frequently Occurring Formula Input Indicators(Top 20)}
    \label{tab:top_formula_indicators}
    \begin{tabular}{r l r}
        \toprule
        Rank & Indicator Name & Frequency \\
        \midrule
        1 & Weight & 88 \\
        2 & Age & 62 \\
        3 & Height [cm] & 28 \\
        4 & Height & 23 \\
        5 & Heart Rate & 17 \\
        6 & Urine Creatinine [mg/dL] & 17 \\
        7 & Tidal Volume & 15 \\
        8 & Blood Creatinine [mg/dL] & 12 \\
        9 & Hemoglobin & 11 \\
        10 & Systolic Blood Pressure & 10 \\
        11 & PaCO\textsubscript{2} & 9 \\
        12 & Cardiac Output & 9 \\
        13 & Respiratory Rate & 9 \\
        14 & PaO\textsubscript{2} & 9 \\
        15 & Serum Creatinine & 8 \\
        16 & Blood Sodium & 8 \\
        17 & Specificity & 7 \\
        18 & Post-dialysis Blood Urea Nitrogen & 7 \\
        19 & Time & 7 \\
        20 & Prevalence & 7 \\
        \bottomrule
    \end{tabular}
\end{table}

\subsection{Model Performance}
\begin{table}[htbp]
\centering
\caption{Performance of LLMs on MedCalc-Eval compared to MedCalc-Bench}
\begin{tabular}{|l|c|c|}
\hline
\textbf{Model} & \textbf{MedCalc-Eval (\%)} & \textbf{MedCalc-Bench (\%)} \\
\hline
DeepSeek-R1-32B & 19.0 & 14.3 \\
\hline
Qwen2.5-7B & 13.1 & 9.7 \\
\hline
Qwen3-8B & 18.2 & 9.6 \\
\hline
Qwen2.5-32B & 25.4 & 21.5 \\
\hline
Qwen3-235B-A22B & 31.1 & 24.5 \\
\hline
Qwen2.5-7B + MedCalc-Env RL & 15.6 & 11.5 \\
\hline
Qwen2.5-32B + MedCalc-Env RL & 40.8 & 20.0 \\
\hline
\end{tabular}
\label{tab:main_results}
\end{table}

\subsection{Ablation Experiments}
\begin{table}[htbp]
\centering
\caption{Ablation study results of Qwen2.5-7B model}
\begin{tabular}{|l|c|c|c|}
\hline
\textbf{Model} & \textbf{MedCalc-Eval (\%)} & \textbf{MedCalc-Bench (\%)} & \textbf{AIME24 (\%)} \\
\hline
Qwen2.5-7B & 13.1 & 9.7 & 16.7 \\
\hline
Qwen2.5-7B + MedCalc-Env RL & 15.6 & 11.5 & 20.0 \\
\hline
\end{tabular}
\label{tab:abalation_results}
\end{table}

\newpage
\section{Prompt templates of MedCalc-Eval \& MedCalc-Env}
\label{detail: prompt}

In this section, we provide the prompt templates\footnote{The original prompts and examples are in Chinese, and we translated them into English for convenience.} in the MedCalc-Eval \& MedCalc-Env, including:
\begin{itemize}
    \item the prompt template of the formula-based question with/without formula explanation;
    \item the prompt template for the scale-based question with/without scale scoring criteria.
\end{itemize}

\subsection{Formula-based Question Prompt}


\begin{tcolorbox}[
    colback=cyan!5,
    colframe=cyan!40!black,
    title=The Prompt for Formula-based Question
]

Patient Information: \placeholder{patient\_info} \\
Please calculate \placeholder{formula\_name}, retain \placeholder{indicator\_precision} decimal places. \\
Let's think step by step and output the final answer within \boxed{xxx:xxx}. For example "\boxed{BMI: 20.5}".

\end{tcolorbox}

\begin{tcolorbox}[
    colback=cyan!5,
    colframe=cyan!40!black,
    title=Formula-based Question Example
]

Patient Information: Red blood cell count 4730.5*10\^9/L, hemoglobin 14.5g/dL. \\
Please calculate mean corpuscular hemoglobin (MCH), retain 2 decimal places. \\
Let's think step by step and output the final answer within \boxed{xxx:xxx}. For example "\boxed{BMI: 20.5}".

\end{tcolorbox}


\subsection{Formula-based Question Prompt with Formula Explanation}


\begin{tcolorbox}[
    colback=cyan!5,
    colframe=cyan!40!black,
    title=The Prompt for Formula-based Question with Formula Explanation
]

\placeholder{formula\_name} \\
Calculation formula: \placeholder{formula} \\
Formula Explanation: \placeholder{formula\_explanation} \\
\\
Patient Information: \placeholder{patient\_info} \\
Please calculate \placeholder{formula\_name}, retain \placeholder{indicator\_precision} decimal places. \\
Let's think step by step and output the final answer within \boxed{xxx:xxx}. For example "\boxed{BMI: 20.5}".

\end{tcolorbox}

\begin{tcolorbox}[
    colback=cyan!5,
    colframe=cyan!40!black,
    title=Formula-based Question Example
]

Glomerular Filtration Rate \\
Calculation formula: 175 * serum creatinine (-1.154) * age (-0.203) * sex
Formula Explanation: Sex: Male: 1, Female: 0.742 \\
Patient Information: Serum creatinine 4.42 mg/dL, age 67, female. \\
Please calculate glomerular filtration rate, retain 3 decimal places. \\
Let's think step by step and output the final answer within \boxed{xxx:xxx}. For example "\boxed{BMI: 20.5}".

\end{tcolorbox}


\subsection{Scale-based Question Prompt}

\begin{tcolorbox}[
    colback=Salmon!5,
    colframe=Salmon!90!Black,
    title=The Prompt for Scale-based Question
]

Patient Information: \placeholder{patient\_info} \\
Please calculate \placeholder{scale\_name}. \\
Let's think step by step and output the final answer within \boxed{xxx:xxx}. For example "\boxed{BMI: 20.5}".

\end{tcolorbox}

\begin{tcolorbox}[
    colback=Salmon!5,
    colframe=Salmon!90!Black,
    title=Scale-based Question Example
]

Patient Information: Limbs/Pelvis: Femoral fracture; comminuted pelvic fracture; knee ligament rupture; Face: LeFort type III fracture (complete facial separation); Abdomen: Small intestine/bladder perforation; subcapsular rupture of liver/spleen; intraperitoneal hemorrhage $\leq$ 1000ml; Chest: Extensive crush injury to the chest; complete aortic transection; Head and neck: Head injury-related headache/dizziness; cervical spine sprain without fracture; minor rupture of external jugular vein (blood loss $\leq 20\%$); thyroid contusion; Body surface: Second- or third-degree burns $\geq 90\%$ of body surface area. \\
Please calculate AIS score. \\
Let's think step by step and output the final answer within \boxed{xxx:xxx}. For example "\boxed{BMI: 20.5}".

\end{tcolorbox}

\subsection{Scale-based Question Prompt with Scoring Criteria}

\begin{tcolorbox}[
    colback=Salmon!5,
    colframe=Salmon!90!Black,
    title=The Prompt for Scale-based Question
]

\placeholder{scale\_name} \\
Scale scoring criteria: \placeholder{scoring\_criteria} \\
\\
Patient Information: \placeholder{patient\_info} \\
Please calculate \placeholder{scale\_name}. \\
Let's think step by step and output the final answer within \boxed{xxx:xxx}. For example "\boxed{BMI: 20.5}".

\end{tcolorbox}

\begin{tcolorbox}[
    colback=Salmon!5,
    colframe=Salmon!90!Black,
    title=Scale-based Question Example
]

Spetsler-Martin grade for an intracranial arteriovenous malformation (AVM) \\
Scale scoring criteria: $[AVM Diameter] [Single Choice]$ \\
$<$3 cm (1 point) ; 3-6 cm (2 points) ; $>$6 cm (3 points) \\
$[AVM Location] [Single Choice]$ \\
Non-functional area (0 point) ; Functional area (1 point) \\
$[AVM Drainage] [Single Choice]$ \\
Superficial venous drainage (0 point) ; Deep venous drainage (1 point) \\
Patient Information: AVM Drainage: Superficial venous drainage, AVM Diameter: 3-6 cm, AVM Location: Non-functional area. \\
Please calculate Spetsler-Martin grade for an intracranial arteriovenous malformation (AVM). \\
Let's think step by step and output the final answer within \boxed{xxx:xxx}. For example "\boxed{BMI: 20.5}".

\end{tcolorbox}
\section{Case Studies}
\label{detail: case}

In this section, we provide the complete model responses\footnote{The original model responses and questions are in Chinese, and we translated them into English for convenience.} for the case studies in Section~\ref{sec: exp}.


\subsection{Case of Scale-based Question}

\begin{tcolorbox}[
   colback=RoyalPurple!5,
   colframe=RoyalPurple!70,
   title=Case study: CURB-65 pneumonia severity score,
   breakable,
   enhanced]
   \begin{tcolorbox}[
      colback=RoyalPurple!5,
      colframe=RoyalPurple!40,
      boxrule=0.3pt,
      left=2pt,
      right=2pt,
      top=1pt,
      bottom=1pt,
      title=Problem]
      Patient Information: Blood Urea Nitrogen: $>$19 mg/dL, Impaired Consciousness: Yes, Age: $<$65 years, Respiratory Rate: $<$30 breaths/min, Blood Pressure: Systolic $<$90 or Diastolic $\leq$60 mmHg. Please calculate the CURB-65 pneumonia severity score. \\
      Let's think step by step and output the final answer within \boxed{xxx:xxx}. For example "\boxed{BMI: 20.5}".
   \end{tcolorbox}

The CURB-65 pneumonia severity scoring system includes the following factors: \\

- C (Confusion, impaired consciousness): 1 point \\
- U (Uremia, blood urea nitrogen $>$20 mg/dL): 1 point \\
- R (Respiratory rate, respiratory rate $\geq$30 breaths/minute): 1 point \\
- B (Blood pressure, systolic blood pressure $<$90 mmHg or diastolic blood pressure $\leq$60 mmHg): 1 point \\
- Age $\geq$65 years: 1 point \\

Based on the information provided in the question, the patient's scoring criteria include: \\
- Impaired consciousness (present): 1 point \\
- Blood urea nitrogen: $>$19 mg/dL (the target blood urea nitrogen is $>$20 mg/dL), but the question does not specify $>$20 mg/dL, only $>$19 mg/dL, so a score of 1 is assumed. (Although it is below 20, the scoring scale is usually not subdivided, so a score of 1 is assumed for this question.) \\
- Respiratory rate: $<$30 breaths/minute (Not meeting the target of 30 beats/minute) \\
- Blood pressure: Systolic pressure $<$90 or diastolic pressure $\leq$60 mmHg (Not meeting the target) \\

The patient does not meet the "age $\geq$65 years" requirement. \\

Comprehensive calculation: \\
1. Impaired consciousness: 1 point \\
2. Blood urea nitrogen: 1 point \\
3. Respiratory rate: 0 point \\
4. Blood pressure: 1 point \\

Total score: 1 + 1 + 0 + 1 = 3 points \\

Final answer:
\(\boxed{3}\)

\end{tcolorbox}
\captionsetup{type=figure}
\captionof{figure}{The complete model response for the scale-based question case study.}
\label{fig: full output for scale based question}

\subsection{Case of Formula-based Question}

\begin{tcolorbox}[
   colback=RoyalPurple!5,
   colframe=RoyalPurple!70,
   title=Case study: Osmol,
   breakable,
   enhanced]

   \begin{tcolorbox}[
      colback=RoyalPurple!5,
      colframe=RoyalPurple!40,
      boxrule=0.3pt,
      left=2pt,
      right=2pt,
      top=1pt,
      bottom=1pt,
      title=Problem]
      Patient information: Glucose 297.0 mg/dL, Blood urea nitrogen 9.44 mg/dL, Sodium 1793.74 mEq/L. Please calculate the estimated osmolality (serum) – osmolality, retain 1 decimal places. \\
      Let's think step by step and output the final answer within \boxed{xxx:xxx}. For example "\boxed{BMI: 20.5}".
   \end{tcolorbox}

To calculate serum osmolality (Osmol), use the formula: \[Osmol = 2[Na^+] + [Glucose]/18 + [BUN]/2.8\], where the units must be consistent. Here, $[Na^+]$ is mEq/L, $[Glucose]$ is mg/dL, and $[BUN]$ is mg/dL. Based on the given data, let's perform the following calculations: \\

1. Glucose ([Glucose] = 297.0 mg/dL) \\
2. Blood urea nitrogen ([BUN] = 9.44 mg/dL) \\
3. Sodium ([Na$^+$] = 1793.74 mEq/L) \\

Substitute the values into the formula and calculate: \\

\[Osmol = 2[Na^+] + \frac{[Glucose]}{18} + \frac{[BUN]}{2.8}\]

\[Osmol = 2(1793.74) + \frac{297.0}{18} + \frac{9.44}{2.8}\]

\[Osmol = 3587.48 + 16.5 + 3.371428571\]

\[Osmol \approx 3607.351428571\]

Take the result to one decimal place: \[Osmol \approx 3607.4\]

Thus, the estimated osmolality (serum) is \(\boxed{3607.4}\).

\end{tcolorbox}
\captionsetup{type=figure}
\captionof{figure}{The complete model response for the formula-based question case study.}
\label{fig: full output for formula based question}

\newpage
\section{Implementation Details of MedCalc-Env}
\label{detail: code}

In this section, we show the implementation details of our \textbf{MedCalc-Env}, a novel reinforcement learning (RL) environment built upon the InternBootcamp framework\footnote{InternBootcamp: https://github.com/InternLM/InternBootcamp}. InternBootcamp is an open-source framework comprising 1000+ domain-diverse task environments specifically designed for LLM reasoning research.

\begin{tcolorbox}[
    colback=OliveGreen!5,
    colframe=OliveGreen!70!Black!70,
    title=Implementation Details of MedCalc-Env
]
\# Create a Bootcamp class, inherit from the Basebootcamp class \\
class MedCalculatorSandbox(Basebootcamp):
    \begin{codebox}{method to generate parameters (construct a single question or validate a response)}
    def _gen_a_case(self, category, name):
        details = self.config[category][name]
        ...
        return {
            "category": category,
            "name": name,
            "inputs": inputs,
            "add_rule": random.random() < self.add_rule_ratio,
            "target": target,
        }

    def case_generator(self):
        category = random.choice(['equation', 'scale'])
        name = random.choice(list(self.config[category].keys()))
        return self._gen_a_case(category, name)
    \end{codebox}

    \begin{codebox}{method to construct the problem statement for a single question}
    def prompt_func(self, case):
        indicators = self.config["indicator"]
        random.shuffle(case["inputs"])
        inp_items = '，'.join(case["inputs"])
        out_item = case["name"]
        ...
        rule = self.rules[case["name"]] if case["add_rule"] else ""
        qes = f"{rule}Patient Information:{inp_items}. Please calculate{out_item}{other_item}."
        return qes + '\nLet's think step by step and output the final answer within \\boxed{xxx:xxx}. For example \\boxed{BMI:20.5}.'
    \end{codebox}

    \begin{codebox}{method to validate whether the answer is correct}
    @classmethod
    def _verify_correction(cls, solution, identity):
        if ':' in solution:
            solution = solution.split(':')[-1].strip()
        elif '：' in solution:
            solution = solution.split('：')[-1].strip()

        return solution.strip() == str(identity['target'])
    \end{codebox}

\end{tcolorbox}

\section{Visual web tool for auditing MedCalc-Eval}
\label{detail: tool}

In this section, we demonstrate a web tool we developed that is convenient for auditing our MedCalc-Eval dataset. It allows for easy previewing of each question in the dataset, error alerts, and statistical results. The interface and functionality are shown below:

\begin{figure}[h!]
  \centering
  \includegraphics[width=0.8\linewidth]{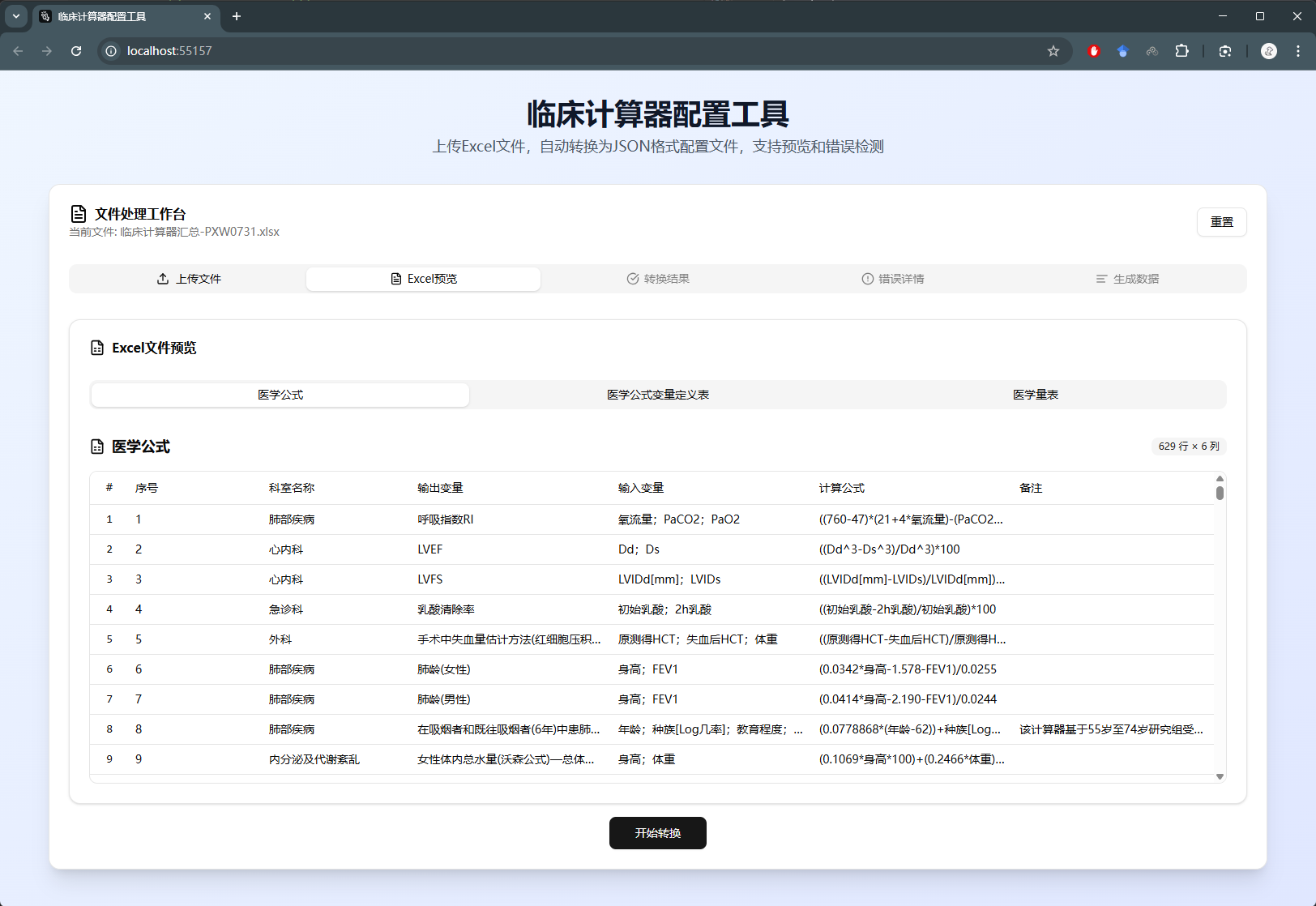}
  \caption{\textbf{Preview page}. This is where you can view detailed data for each class of question.}
  \vspace{8pt}

  \includegraphics[width=0.8\linewidth]{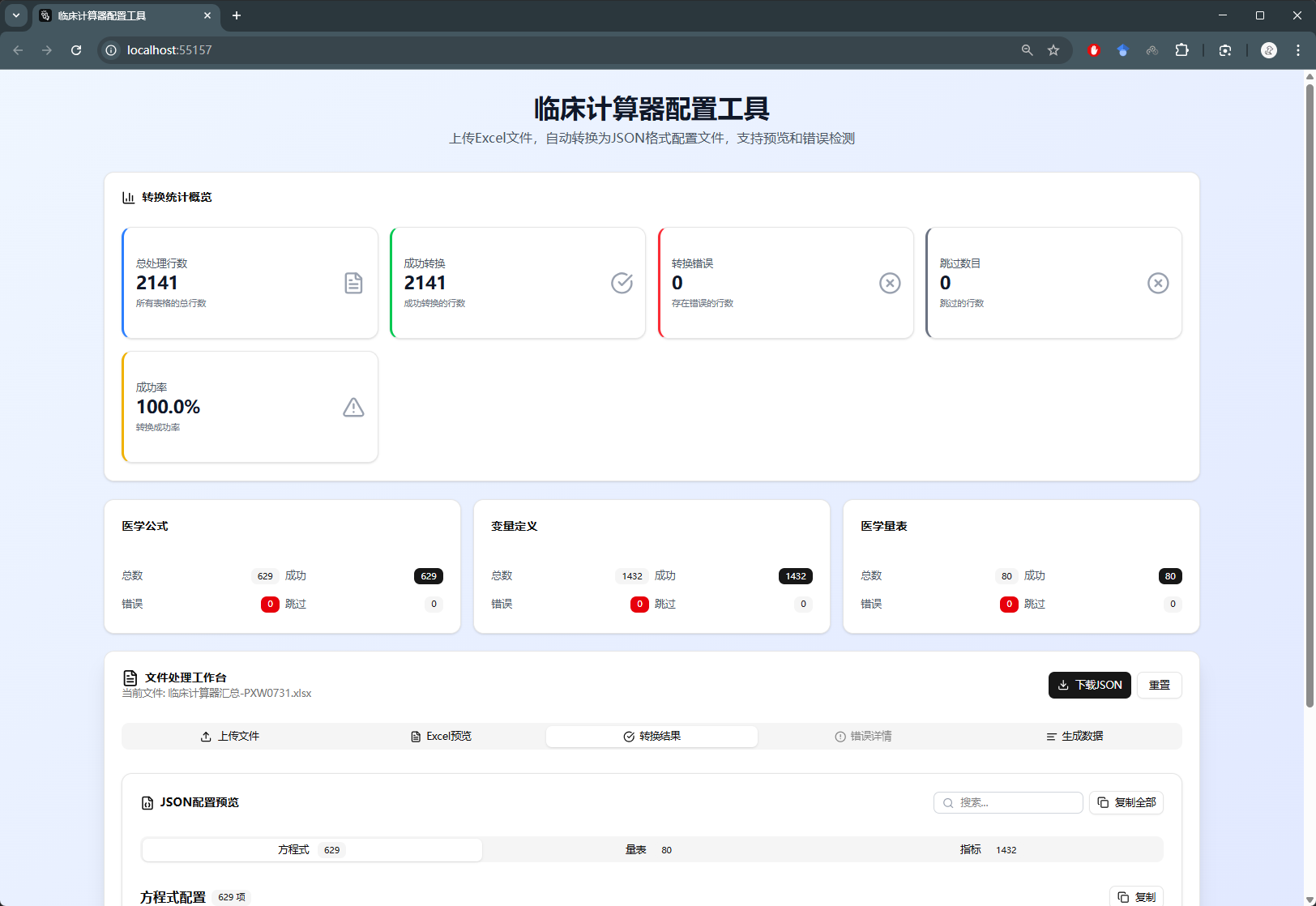}
  \caption{\textbf{Statistical results}. This is used to view the overall statistics of the dataset, including the type and number of questions, and the number of errors.}
  \vspace{-8pt}
\end{figure}

\begin{figure}
  \centering
  \includegraphics[width=0.8\linewidth]{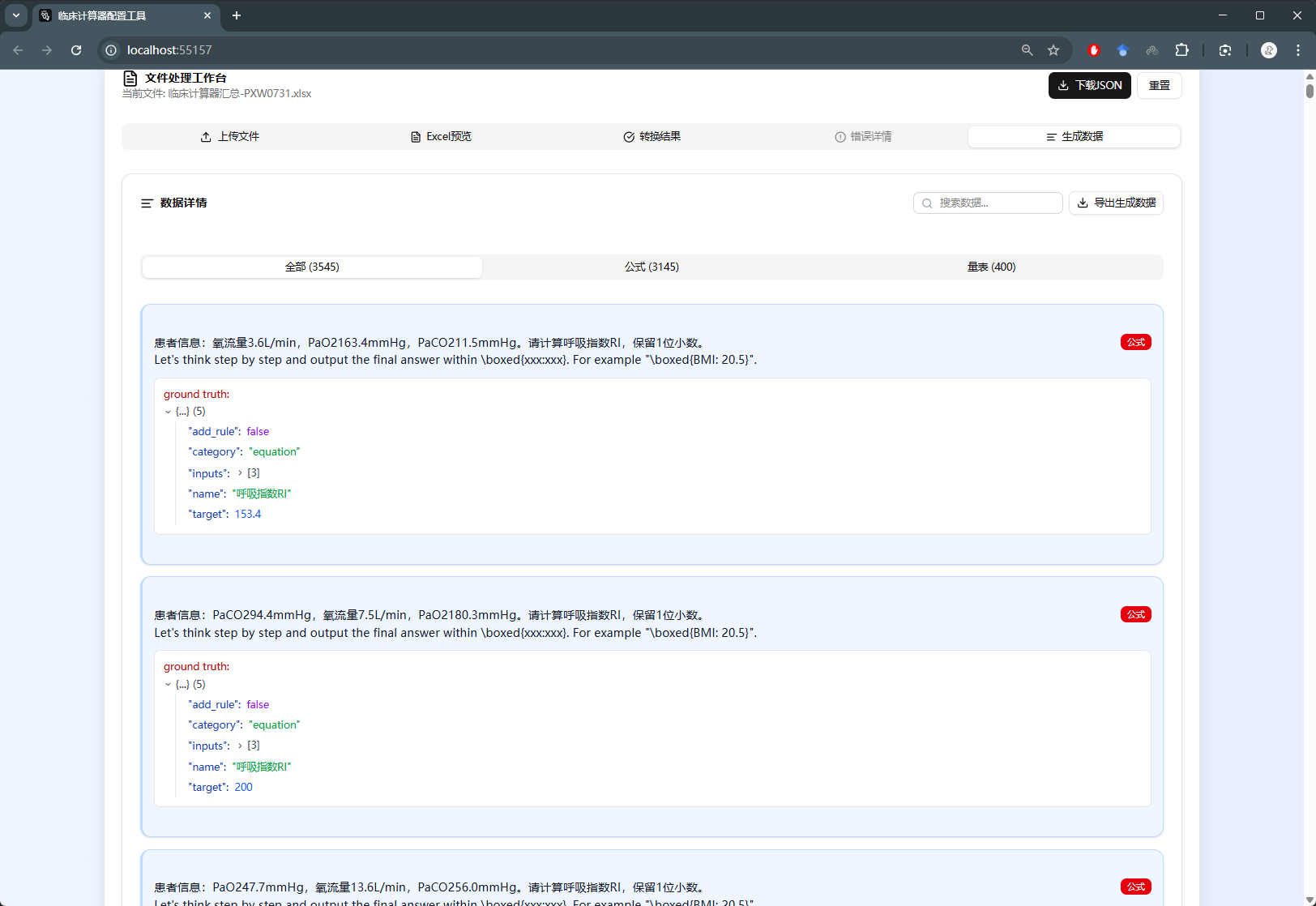}
  \caption{\textbf{Generated data}. This is used to preview and review the generated data, and indicate any potential errors.}
  \vspace{-8pt}
\end{figure}

\end{document}